\documentclass[a4paper,UKenglish,cleveref, autoref, thm-restate]{lipics-v2019}
\usepackage{tikz}
\usepackage{pgf-pie}
\usepackage{caption,stackengine}
\usepackage{tabularx}
\usepackage{array}
\usepackage{adjustbox}
\usepackage{algorithm2e}
\usetikzlibrary{automata,positioning}

\usepackage[colorinlistoftodos,prependcaption,textsize=tiny]{todonotes}

\title{Comparative study of multi-person tracking methods} 

\titlerunning{Comparative study of multi-person tracking methods} 

\author{Denis Mbey Akola}{Institut Polytechnique de Paris, France}{denis.akola@ip-paris.fr}{}{}

\authorrunning{D. Akola} 

\Copyright{D. Akola} 
\ArticleNo{1}


\category{} 

\relatedversion{} 

\supplement{}

\nolinenumbers 

\begin{document}

\maketitle

\begin{abstract}
This paper presents a study of two tracking algorithms (SORT and Tracktor++~) that were ranked first positions on the MOT Challenge leaderboard~\footnote{The MOTChallenge web page: https://motchallenge.net}. The purpose of this study is to discover the techniques used and to provide useful insights about these algorithms in the tracking pipeline that could improve the performance of MOT tracking algorithms. To this end, we adopted the popular tracking-by-detection approach. We trained our own Pedestrian Detection model using the MOT17Det dataset. We also used a re-identification  model trained on the MOT17 dataset for Tracktor++  to reduce the false re-identification alarms. We then present experimental results  that show that Tracktor++ is a better multi-person tracking algorithm than SORT. We also performed ablation studies to discover the contribution of the re-identification(RE-ID) network and motion to the results of Tracktor++. We finally conclude by providing some recommendations for future research.
\end{abstract}

\section{Introduction}
\label{sec:typesetting-summary}
 A large amount of video data is being produced today. This is largely due to the recent improvement in video technology that makes it much easier for people to record high-quality videos. Videos contain rich information which could be very useful when we are able to understand these pieces of information using various analytical tools. Thus, being able to analyze videos has a wide range of applications in surveillance, autonomous driving,  and pedestrian tracking.
Person tracking  in video scenes is still an active challenge, especially in crowded environments. This is a result of the occlusions and fast detection that often occur in such settings. 

The person-tracking task is divided into two main aspects. First, we have to be able to detect people accurately in  a video frame, and we must be able to establish some relationships that connect people in one video frame to another. This involves being able to identify each person and retain their identities in future video frames.
In the detection step, object detection models are used to detect the objects of interest. In our case, the people in the videos. The data association  is the matching step where persons in video frames are linked to form a trajectory over time~\cite{2019}.

With the recent advances in object detection, we have robust object detectors and these perform fairly well, but they still struggle to make good predictions in crowded, partial, and fully occluded scenes.
Data association or tracking is still a challenge because of false alarms and missing data, especially in crowded environments. Moreover, data association is a challenge because of the complex and dynamic movement activity of human beings.

In this paper, we present a comparative study of two  tracking algorithms and provide some recommendations for future research in multiple-object tracking

\section{Related work}
Multiple-object tracking finds applications in many areas ranging from autonomous driving to surveillance~\cite{article}. Despite the wide application domain of tracking, it is still a challenge that is actively being investigated by the research community. From a broader point of view, there are basically two approaches to the  multi-object tracking problem. These two are detection-based tracking and detection-free tracking~\cite{Luo2014MultipleOT}.

In the detection-based tracking approach, objects of interest are first detected using classical deep-learning object detection models, and the tracks are subsequently associated with forming  trajectories. This approach is promising as it is best suitable for online-tracking applications like autonomous driving~\cite{https://doi.org/10.48550/arxiv.1411.7935}. The second kind of approach is detection-free tracking  which requires manual initialisation of a fixed number of objects in the first frame. Detection-free tracking methods then try to localize these objects in subsequent frames. This method might not work well  because the objects of interest appearances may change drastically over time making it difficult for the target object to be tracked. Also, little information about the target object is always available at the beginning of the tracking process. Mostly the information available as input to the tracking algorithm is the initial bounding box  which distinguishes the object of interest from the background~\cite{Zhang_2013_CVPR}. Since we are interested in tracking people in videos, for the sake of terminology, we will proceed to call such a task multi-person tracking.
In the remainder of this section, we provide a brief review of related work on object detectors for multi-person tracking and some of the state-of-the-art object tracking methods.
\subsection{Object detectors for multi-person tracking}
As identified in the detection-based tracking approach, object detection models are used for the object detection task. Object detection involves scanning and searching for objects of certain classes (e.g. humans, cars, and buildings) in an image/video frame~\cite{article}.  Object detection is at the heart of scene understanding~\cite{9578417}. Thus, the quality of detections from object detection models affects the results we can obtain in a typical tracking task. The most common object detection models used in the detection-based tracking are R-CNN~\cite{DBLP:journals/corr/GirshickDDM13}, Fast R-CNN~\cite{DBLP:journals/corr/Girshick15}, Faster R-CNN~\cite{DBLP:journals/corr/RenHG015}, and Mask R-CNN~\cite{https://doi.org/10.48550/arxiv.1703.06870}. These algorithms are inspired by convolutional neural networks which are well adapted to working with images.\\
R-CNN network has three modules~\cite{DBLP:journals/corr/GirshickDDM13}. The first module embeds the region proposal network. This module generates category-independent region proposals. The second module contains a large convolutional neural network that generates a fixed-length feature vector from the regions proposed by the region proposal network.
The third module contains a set of class-specific linear Support Vector Machines (SVMs) that are used to classify input images into classes. The region proposal network uses a selective search algorithm to propose approximate object regions in the input image ~\cite{uijlings2013selective}.
R-CNN network is quite computationally expensive  because both the region proposal and classification tasks are performed separately without computation sharing.\\ 
Fast R-CNN~\cite{DBLP:journals/corr/Girshick15} was engineered to overcome the challenges of R-CNN notable for the run-time problem of R-CNN. Fast R-CNN takes an image as input and then processes the whole image with several convolutional (conv) and max pooling layers to produce a convolutional feature map corresponding to the image. Fast R-CNN has the region of interest (IOU) pooling layer which extracts fixed-length feature vectors from the feature map. Each of the feature vectors generated from the ROI layer are fed into a sequence of fully connected layers that finally branch out into sibling output layers(the softmax layer and a layer that contains the bounding box for the predicted class in the softmax layer). However, Fast R-CNN uses a selective search method to generate the pooling map from the convolution feature map which slows down its operation~\cite{DBLP:journals/corr/Girshick15}~\cite{article}.\\
Faster R-CNN~\cite{DBLP:journals/corr/RenHG015} has two modules. The first  module is a deep fully convolutional network layer that proposes regions and the second module is the  Fast R-CNN detector. Faster R-CNN networks adopt the fast R-CNN network and make improvements to the way the region proposals are generated in Fast R-CNN to reduce the overall run-time of the Faster R-CNN network. The region proposal network (RPN) in Faster R-CNN takes an image as input and outputs rectangular object proposals each with an objectness score. The intuition behind Faster R-CNN is sharing computations between the Region proposal network and the R-CNN detector. As a result, a formal assumption is that both the RPN and R-CNN networks share the same convolutional layers. Anchors are placed at each convolution. Anchors (spatial windows of different sizes and different aspect ratios) are placed at different locations in the input feature map. Faster R-CNN as the name suggests has a high detection speed than its counterparts~\cite{DBLP:journals/corr/RenHG015}.\\
Mask R-CNN~\cite{https://doi.org/10.48550/arxiv.1703.06870} is yet another extension of Faster R-CNN for both object detection and segmentation. The work of ~\cite{article} provides an extensive review of object detectors.
For this paper, we trained a Faster R-CNN with ResNet-50~\cite{simonyan2014very} and Feature Pyramid Networks (FPN)~\cite{https://doi.org/10.48550/arxiv.1612.03144}  on the MOT17Det~\footnote{MOT17Det: https://motchallenge.net/data/MOT17Det/} pedestrian detection dataset.

\subsection{Data association methods}
Data association involves computing the similarities between tracklets and detection boxes and then matching them using various similarity criteria. Different metrics have been employed to compute these similarities. Likewise, different strategies have been used to match tracklets and detection boxes. In this section, we will offer a brief review of some of the commonly used methods to compute the similarity as well as the various matching strategies in multi-object tracking. The common similarity metrics used include motion, appearance,  and distance/location.\\
For instance, in the SORT tracking algorithm~\cite{7533003}, the authors used both motion and location similarity metrics in a simple way. First, they used Kalman Filter~\cite{Klmn1960ANA} to predict the location of tracks in new frames and used Intersection over Union (IOU) to compute the similarity score between detection boxes and predicted boxes. The Hungarian  algorithm is used to assign detections to existing targets. Also, a minimum IOU was imposed to reject assignments where the detection to target overlaps is less than a certain threshold. \\
The authors of CenterTrack~\cite{DBLP:journals/corr/abs-2004-01177} designed their tracktor to learn object motion and achieve robust results in case of large camera motions or low frame rate. In their approach, their tracking network was able to learn and match prediction boxes anywhere in the receptive field even if there existed no overlap between the boxes. For the motion model, they used a sparse optical flow estimation network which was learned together with the detection network and does not require dense supervision. A greedy assignment strategy was used in their matching step.
In TransTrack~\cite{DBLP:journals/corr/abs-2012-15460}, the authors used a joint-detection and tracking pipeline to achieve detection and tracking in a single stage. They applied object features from previous frames as queries of current frames and as a result, introduced a set of object queries for detecting new incoming objects. Kalman filter was used in their motion step. Also, they included a Re-identification (Re-ID) model to help them re-identify objects after they go occluded for a long time using the appearance similarity. For the matching step, they used the classic Hungarian algorithm and Non-Maximum Suppression(NMS) merging method inspired by ~\cite{DBLP:journals/corr/abs-2101-02702}.\\
In Tracktor++~\cite{2019}, the authors used the bounding box regression of an object detector to predict the position of an object in the next frame. In this work, they adopted two motion models to handle bounding box positions in future frames. In the case of sequences with moving cameras, they applied the camera motion compensation(CMC) technique by aligning frames through registration using an enhanced  Correlation Coefficient as introduced in ~\cite{evangelidis:hal-00864385}. For low frame sequences, they applied the constant velocity assumptions (CVA) ~\cite{Andriyenko2011MultitargetTB} for all object frames. They also used a Re-ID model based on appearance vectors generated using a Siamese neural network. For the matching strategy, they used the Hungarian algorithm.\\
In DeepSort~\cite{7533003}, the authors improved the SORT tracking algorithm by adopting a Re-ID network to improve the tracking performance of SORT. To this end, they trained a re-identification CNN network to extract appearance features from detection boxes. Likewise, they adopted a new matching strategy which first matches detection boxes to the most recent tracklets, and then to the lost tracks. 

Following this review, we now offer comparative studies on the performance of SORT and Tracktor++ tracking algorithms in the MOT17 dataset~\cite{article_2016} and provide some insights for the development of future multi-tracking algorithms.

\section{Object Detection Model}
\subsection{Pedestrain Detection Model}
In this comparative work, we adopted Faster R-CNN network  with  ResNet-50~\cite{simonyan2014very} and Feature Pyramid Networks(FPN)~\cite{https://doi.org/10.48550/arxiv.1612.03144} trained on the MOT17Det dataset~\footnote{MOT17Det: https://motchallenge.net/data/MOT17Det/} for pedestrian detection. The Faster R-CNN model has a three-stage architecture. In the first stage, the Region Proposal Network (RPN) was used to generate region proposals for each image. However, the RPN network was adapted by replacing the single-scale feature map with FPN as discussed in ~\cite{DBLP:journals/corr/LinDGHHB16} and implemented in ~\cite{2019}.
The second stage is the Fast R-CNN network was used to extract features from the region proposals generated by the RPN stage. The last layer contains classification and regression heads. The classification head assigns an objectness score to each box proposal by evaluating the likelihood of a proposal region showing a pedestrian. The regression head has the responsibility of refining the bounding box location tightly around an object. The final set of object detections was obtained by applying non-maximum suppression to the refined bounding box proposals.

\subsection{Model implementation and training}
\begin{figure}[!htb]
  \centering
  \includegraphics[scale=0.90]{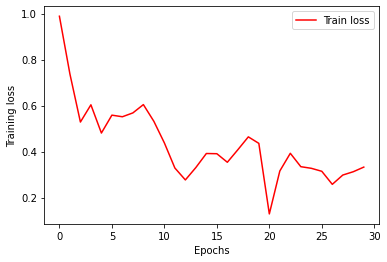}
  \caption{Training loss curve}
  \label{fig3:dif_config_2}%
\end{figure}

\begin{figure}[!htb]
  \centering
  \includegraphics[scale=0.45]{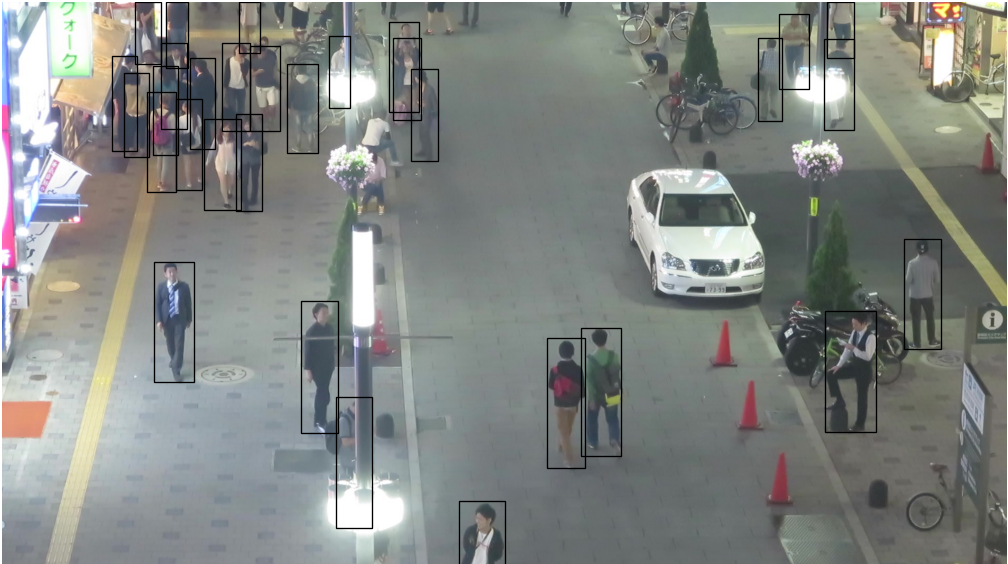}
  \caption{Object detector output}
  \label{image}%
\end{figure}

The Pedestrian Detection model was implemented in Pytorch using Faster R-CNN multi-object detector with Feature Pyramid Network (FPN) with ResNet-50~\cite{simonyan2014very} as the feature extractor. Moreover, we replaced the Region of Interest (ROI) with crop and resize pooling layer as in the tracktor++ paper~\cite{2019}. The model was trained on a single NVIDIA GeForce GTXTITAN X GPU with 12 GB of memory for 12 hours. The object detector layers  parameters were updated with an initial learning rate of 0.001, batch size of 8, momentum of 0.9, and weight decay of 0.0005. We used the Stochastic Gradient Descent(SGD)~\cite{ruder2016overview} optimizer to update the learning rate if it remained constant for 10 training epochs. The training loss curve of the model is shown in figure ~\ref{fig3:dif_config_2}. The model achieved an average precision (AP) score of 0.815 and an average recall(AR) of 0.852.
Figure ~\ref{image} shows the bounding boxes of an image that was processed by the object detection model.
\section{SORT Tracking}
The SORT tracking algorithm was introduced by Alex Bewley et al in 2017~\cite{7533003}. SORT makes the assumption that the presence of short-term and long-term occlusions occur less frequently and thus, ignores these edge cases. The SORT algorithm tried to alleviate the problem of computation complexity and therefore it does not use a  re-identification model in the tracking pipeline as well.
Kalman Filter and Hungarian algorithms were employed in SORT to handle motion prediction and data-matching respectively.  SORT propagates target identities into next frames by using the inter-frame displacements of each object with a linear constant velocity model which was independent of other objects and camera motion. SORT created new tracks only if the Intersection Over Union(IOU) of new detections was less than a certain threshold ($IOU_{min}$). When this condition is met, SORT creates new tracklets based on the geometry of detection bounding boxes. SORT then initialized the velocity of new tracks to zero because the velocity of tracks was unobserved initially.
 SORT associates new tracklets with detections to accumulate enough evidence to prevent tracking of false positives. SORT  deletes tracklets if they were not detected for a number of frames ($T_{Lost}$). The value of $T_{Lost}$ was set to 1.

\section{Tracktor++ tracking}
The Tracktor++ tracking algorithm proposed by the authors of ~\cite{2019} is an online tracking algorithm that exploits the regression head of object detection models to regress and classify bounding boxes for the purpose of multi-object tracking. In fact, the authors argued that all we need to perform multi-object tracking is an object detection model. The challenge of multi-object tracking is to  form a trajectory using frames from a video sequence. A typical trajectory of frames is defined as a list of ordered object bounding boxes $T_{k} =\{b_{t_{1}}^{k},b_{t_{2}}^{k},... \}$ where $b_{t_{1}}^{k},b_{t_{2}}^{k},... $ denotes bounding boxes with coordinates $(x,y,w,h)$ and $t$ represents a frame in the video.
In the initialization step of Tracktor++, tracks are formed using the initial set of detections from the object detection model. The regression head of the pedestrian detection model was then leveraged to regress active trajectories to the current frame, $t$. To this end, the bounding boxes of objects in the frame, $t-1$ are regressed to the new position of the objects in the frame,$t$. For newer detections following the initial detections, a new trajectory was started only if the IOU with any of the existing trajectories was less than a threshold, ${\lambda_{new}}$. In some cases, the two frames might have different tracked objects. In such situations, Tracktor++ utilized two well-known motion models to improve the bounding box positions in future frames. Camera Motion Compensation (CMC) was used for sequences with a moving camera  to align the frame using the Enhanced Correlation Coefficient(ECC). Otherwise, for low frame sequences, a constant velocity assumption(CVA) was applied. Tracker also used a short-term re-identification (reID) based on appearance vectors generated by a Siamese neural network. This network helped in re-matching deactivated tracklets with new bounding boxes based on the IOU threshold.

\section{Tracking metrics}
In this section, we will explain the MOT tracking performance metrics that would be used in our comparative study for the tracking results of SORT and Tracktor++. First and foremost, there is the need for standard MOT performance metrics so that the performance of various tracking algorithms can be verified. Using such benchmark performance metrics would help the research community to learn insights from models that have better performance and this would positively influence future research in this domain. Thus, we adopt the MOT metrics defined in ~\cite{10.1155/2008/246309}  and ~\cite{Ju:17}. The evaluation metrics for tracking are: global min-cost F1 score (IDF1, $(\uparrow)$)~\cite{Ju:17}, Multi-object tracking accuracy (MOTA,  $(\uparrow)$)~\cite{10.1155/2008/246309}, Recall (Rcll, $(\uparrow)$)~\cite{Ju:17}, number of mostly tracked trajectories (MT,$(\uparrow)$)~\cite{Ju:17}, number of mostly lost trajectories(ML, $(\downarrow)$)~\cite{Ju:17}, number of false detections (FP,($\downarrow$))~\cite{Ju:17}, number of missed detections (FN, ($\downarrow$))~\cite{Ju:17}, and ID switches( ID Sw., $(\downarrow)$)~\cite{Ju:17} which is number of times an ID switches to a different
previously tracked object. Evaluation metrics  with ($\uparrow$), mean higher scores denote better performance; while for metrics with ($\downarrow$), mean lower scores denote better performance. 
We performed tracking experiments using the MOT17 dataset. We used the training data for this experiment because the training set had ground truth detection which we could use to compute the motmetrics for both SORT and Tracktor++. The MOT challenge datasets consist of several challenging pedestrian tracking sequences with crowded scenes and occlusions.
\subsection{Experiments with SORT algorithm}
In this section, we present the tracking results of using the SORT algorithm on the MOT17 training dataset.
We used private detections  to evaluate the performance of SORT. By private detections,  we mean we used our object detection model to predict the bounding boxes for each frame  in the training set. The tracking results for SORT are shown in Table ~\ref{sort-tracking}.

\begin{table}[!htb]

    \centering
    \begin{adjustbox}{width=1.1\textwidth}

 \begin{tabular} {c|c|c|c|c|c|c|c|c|c|c|}
  Sequence & MOTA$\uparrow$ (\%)  & IDF1$\uparrow$ &Rcll$\uparrow$  &Prcn$\uparrow$   &MT$\uparrow$  &ML$\downarrow$   & FP$\downarrow$& FN$\downarrow$& ID Sw.$\downarrow$&Time $\downarrow$\\\hline

MOT17-02-DPM &11.60 &0.50&1.30&59.40&0&62&159&18348&183&161.18\\
MOT17-02-FRCNN& 40.90&43.60&51.40&96.80&12&14&313&9024&149&159.69\\
MOT17-02-SDP&10.60  &0.50&1.30&59.40&0&62&159&18348&183&159.68\\
MOT17-04-DPM & 20.00 &0.20&0.20&61.20&0&83&59&47464&56&291.48\\
MOT17-04-FRCNN & 76.00 &74.70&77.60&98.20&43&10&667&10676&71&291.56\\
MOT17-04-SDP& 31.00   &0.20&0.20&61.20&0&83&59&47464&56&291.56\\
MOT17-05-DPM & 16.20 &2.30&2.70&72.60&0&133&71&6729&105&161.84\\
MOT17-05-FRCNN & 47.80 &50.70&51.90&96.10&9&44&145&3325&139&161.26\\
MOT17-05-SDP & 17.20 &2.30&2.70&72.60&0&133&71&6729&105&161.87\\
MOT17-09-DPM& 19.20  &0.70&1.00&84.40&0&26&10&5271&35&173.74\\
MOT17-09-FRCNN & 64.10 &55.60&66.10&98.40&12&1&58&1805&47& 138.78\\
MOT17-09-SDP & 19.60  &0.70&1.00&84.40&0&26&10&5271&35&138.65\\\
MOT17-10-DPM &18.70  &0.80&3.90&65.30&0&57&267&12337&448&173.74\\
MOT17-10-FRCNN& 54.00 &26.50&60.30&96.30&9&5&294&5096&518& 174.24\\
MOT17-10-SDP & 22.70 &0.80&3.90&65.30&0&57&267&12337&448&173.51\\
MOT17-11-DPM & 24.30   &1.00&1.40&65.00&0&75&71&9304&86&235.41\\
MOT17-11-FRCNN & 69.30 &58.90&72.40&97.40&23&7&179&2609&110&234.89\\
MOT17-11-SDP & 19.30 &1.00&1.40&65.00&0&75&71&9304&86& 235.23\\
MOT17-13-DPM &23.00 &1.40&3.00&52.70&0&110&318&11288&267&201.77 \\
MOT17-13-FRCNN &50.10 &47.10&54.70&96.10&18&26&262&5271&273& 201.74\\
MOT17-13-SDP &17.00 &1.40&3.00&52.70&0&110&318&11288&267&201.69\\ \hline
\textbf{OVERALL} &\textbf{32.03}&\textbf{26.70}&\textbf{23.00}&\textbf{95.30}&\textbf{126}&\textbf{1199}&\textbf{3828}&\textbf{259288}&\textbf{667}&\textbf{4087.72}\\

\end{tabular}
\end{adjustbox}
    \caption{SORT tracking results for MOT17 for private detections }
    \label{sort-tracking}
\end{table}

\subsection{Experiment with Tracktor++  algorithm}
Tracktor++ utilizes the Re-ID model to recover tracks after they have been occluded for some time. To this end, we trained the Re-ID model with ResNet-50~\cite{simonyan2014very} as the feature extractor using the MOT17 Dataset. The Re-ID model  had a batch size of 32 and a learning rate of 0.0001. AdaGrad optimization algorithm~\cite{ruder2016overview} was used as our optimizer. The model was trained on a single NVIDIA GeForce GTXTITAN X GPU with 12 GB of memory for about 6 hours.
The tables ~\ref{tracktor++1},~\ref{tracktor++2} and ~\ref{tracktor+++3} shows tracking results of Tracktor++ for all private detections on the MOT17 Dataset~\footnote{MOT17Dataset : https://motchallenge.net/data/MOT17/}. The MOT17 Challenge~\footnote{The MOTChallenge web page: https://motchallenge.net} also provides detections for the dataset that was generated using classical object detectors like Faster RCNN~\cite{DBLP:journals/corr/Girshick15}, Deformable Parts Model~\cite{felzenszwalb2010object}, and Scaled Dependent Pooling(SDP)~\cite{7780603} detector.

\begin{table}[!htb]

    \centering
    \begin{adjustbox}{width=1.1\textwidth}

   \begin{tabular} {c|c|c|c|c|c|c|c|c|c|c|}
  Sequence & MOTA$\uparrow$ (\%) & IDF1$\uparrow$ &Rcll$\uparrow$  &Prcn$\uparrow$   &MT$\uparrow$  &ML$\downarrow$   & FP$\downarrow$& FN$\downarrow$& ID Sw.$\downarrow$& Time $\downarrow$  \\\hline
   MOT17-02-DPM& 46.70 &42.30&50.40&95.50&11&13&439&9215&245&219.04 \\
MOT17-02-FRCNN& 46.70 &42.30&50.40&95.50&11&13&439&9215&245&220.21\\
MOT17-02-SDP &46.70 &42.30&50.40&95.50&11&13&439&9215&245&220.34\\
MOT17-04-DPM & 75.30 &69.80&76.10&99.50&44&16&196&11350&200&407.58\\
MOT17-04-FRCNN &75.30 &69.80&76.10&99.50&44&16&196&11350&200&407.57\\
MOT17-04-SDP &75.30 &69.80&76.10&99.50&44&16&196&11350&200&406.08\\
MOT17-05-DPM & 58.50 &55.20&65.60&96.60&48&14&162&2376&330&225.54\\
MOT17-05-FRCNN & 58.50 &55.20&65.60&96.60&48&14&162&2376&330&227.41\\
MOT17-05-SDP & 58.50 &55.20&65.60&96.60&48&14&162&2376&330& 226.99\\
MOT17-09-DPM & 66.10 &55.10&67.80&99.30&12&2&27&1715&63&186.42\\
MOT17-09-FRCNN &66.10 &55.10&67.80&99.30&12&2&27&1715&63& 187.13\\
MOT17-09-SDP & 66.10 &55.10&67.80&99.30&12&2&27&1715&63& 186.69 \\
MOT17-10-DPM & 61.40 &36.00&78.90&89.00&34&1&1251&2707&1001& 246.05\\
MOT17-10-FRCNN &61.40 &36.00&78.90&89.00&34&1&1251&2707&1001&  246.99 \\
MOT17-10-SDP&61.40 &36.00&78.90&89.00&34&1&1251&2707&1001& 247.21\\
MOT17-11-DPM &74.10 &62.60&77.90&97.80&38&7&168&2081&194&321.95\\
MOT17-11-FRCNN &74.10 &62.60&77.90&97.80&38&7&168&2081&194&322.29\\
MOT17-11-SDP & 74.10 &62.60&77.90&97.80&38&7&168&2081&194&320.05\\
MOT17-13-DPM & 51.00 &37.10&83.20&87.40&73&5&1399&1956&2353&292.01\\
MOT17-13-FRCNN & 51.00 &37.10&83.20&87.40&73&5&1399&1956&2353&292.37\\
MOT17-13-SDP& 51.00 &37.10&83.20&87.40&73&5&1399&1956&2353&292.69\\ \hline
\textbf{OVERALL}& \textbf{64.90}&\textbf{55.70}&7\textbf{2.00}&\textbf{95.70}&\textbf{780}&\textbf{174}&\textbf{10926}&\textbf{94200}&\textbf{13158}&\textbf{5702.61}\\
\end{tabular}
\end{adjustbox}
    \caption{Tracktor++ with no REID network, and no motion model tracking results for MOT17 for private detection without motion model}
    \label{tracktor++1}
\end{table}

\begin{table}[!htb]

    \centering
    \begin{adjustbox}{width=1.1\textwidth}

   \begin{tabular} {c|c|c|c|c|c|c|c|c|c|c|}
  Sequence & MOTA$\uparrow$ (\%)  & IDF1$\uparrow$ &Rcll$\uparrow$  &Prcn$\uparrow$   &MT$\uparrow$  &ML$\downarrow$   & FP$\downarrow$& FN$\downarrow$& ID Sw.$\downarrow$&Time $\downarrow$\\\hline
    MOT17-02-DPM&48.40 &43.90 &53.40 &94.90 &14 &13 &500 &661 &210&284.64 \\
MOT17-02-FRCNN & 47.70  &43.60 &53.40 &95.50 &14 &13 &602 &653 &213&282.70\\
MOT17-02-SDP& 49.20  &44.80 &53.50 &95.50 &12 &11 &450 &637 &223&283.75 \\
MOT17-04-DPM  &76.90  &70.20 &77.10 &96.10 &44 &15 &479 &10871 &77&598.75\\
MOT17-04-FRCNN  &75.10  &70.30 &77.20 &96.40 &47 &14 &300 &10841 &73&602.42\\
MOT17-04-SDP  &76.70  &70.10 &77.20 &96.60 &44 &15 &290 &10834 &73&600.13\\
MOT17-05-DPM &58.70   &55.60 &72.90 &94.60 &59 &15 &718 &1876 &91&282.24\\
MOT17-05-FRCNN &58.79 &56.40 &72.80 &94.90 &58 &13 &684 &1879 &85&269.82 \\
MOT17-05-SDP&59.40   &55.70 &72.60 &94.50 &56 &15 &216 &1894 &96&265.59 \\
MOT17-09-DPM  &66.50  &55.25 &73.80 &90.30 &15 &2 &300 &1395 &46&222.55\\
MOT17-09-FRCNN &66.80  &55.20 &73.40 &92.70 &14 &2 &401 &1416 &47& 221.47\\
MOT17-09-SDP&67.70   &55.80 &74.20 &91.70 &14 &2 &200&1372 &44 & 221.79\\
MOT17-10-DPM &62.25  &49.80 &82.70 &94.40 &37 &1 &162 &2219 &253&315.17 \\
MOT17-10-FRCNN &61.70  &49.30 &82.40 &89.90 &37 &1 &741 &2264 &267& 319.58\\
MOT17-10-SDP   &67.28&49.20 &83.00 &94.30 &39 &1 &679 &2181 &273& 329.58\\
MOT17-11-DPM  &75.65  &65.50 &80.90 &88.60 &44 &6 &391 &1805 &47& 319.78\\
MOT17-11-FRCNN&75.95  &63.00 &81.20 &87.60 &44 &6 &399 &1772 &45 &390.09\\
MOT17-11-SDP&75.76   &62.78 &80.90 &84.60 &43 &6 &385 &1799 &48 &396.82\\
MOT17-13-DPM  &52.52   &38.70 &86.70 &95.90 &80 &4&2768 &1553 &372&348.63\\
MOT17-13-FRCNN &53.81&38.50 &86.70 &86.70 &81 &3 &2386 &1552 &376& 346.40  \\
MOT17-13-SDP  &51.65  &38.20 &86.40 &87.00 &82 &3 &1517 &1589 &362&347.23\\ \hline
\textbf{OVERALL}&\textbf{65.77}  &\textbf{56.23} &\textbf{94.8 }&\textbf{84.7 }&\textbf{878} &\textbf{161} &\textbf{14568}&\textbf{85063 }&\textbf{3321}&\textbf{7291.05} \\
\end{tabular}
\end{adjustbox}
    \caption{Tracktor++ with REID network and no motion model tracking results for MOT17 for private detection without motion model}
    \label{tracktor++2}
\end{table}

\begin{table}[!htb]

    \centering
    \begin{adjustbox}{width=1.1\textwidth}

  \begin{tabular} {c|c|c|c|c|c|c|c|c|c|c|}
  Sequence & MOTA$\uparrow$ (\%) & IDF1$\uparrow (\%)$ &Rcll$\uparrow$  &Prcn$\uparrow$   &MT$\uparrow$  &ML$\downarrow$   & FP$\downarrow$& FN$\downarrow$& ID Sw.$\downarrow$ &Time $\downarrow$ \\\hline
  MOT17-02-DPM& 50.00 &44.90&54.30&85.40&14&13&490&630&225&522.83\\
MOT17-02-FRCNN & 52.40 &43.80&54.90&85.20&14&13&590&558&238&520.16\\
MOT17-02-SDP&52.60&45.20&53.20&85.90&14&12&420&432&214&524.97\\
MOT17-04-DPM&77.80 &70.90&77.60&96.10&44&15&471&450&81&1114.41\\
MOT17-04-FRCNN&77.90 &69.90&77.50&96.40&44&15&286&360&79&1110.55\\
MOT17-04-SDP& 77.00 &73.00&77.70&96.30&44&15&280&250&76&1118.18\\
MOT17-05-DPM& 60.20 &54.40&74.10&77.80&58&13&710&433&92&368.38 \\
MOT17-05-FRCNN &59.60 &56.80&78.60&77.52&56&13&256&1861&88&368.96\\
MOT17-05-SDP& 60.70 &56.60&74.00&78.49&56&13&207&1739&93& 369.63\\
MOT17-09-DPM & 66.20 &52.10&73.20&90.98&15&2&384&1380&45&458.14\\
MOT17-09-FRCNN&66.70 &53.10&73.40&91.55&14&2&365&1399&40&459.20\\
MOT17-09-SDP&65.80&59.60&73.80&91.65&14&2&355&151&45&460.08 \\
MOT17-10-DPM& 64.60 &50.70&83.00&77.96&36&1&165&2015&278& 752.58\\
MOT17-10-FRCNN &63.90 &50.50&83.50&78.59&36&1&673&2209&286&757.78\\
MOT17-10-SDP &64.70 &50.20&84.00&78.90&36&1&695&2010&293&752.84\\
MOT17-11-DPM &77.10 &66.20&87.60&86.70&42&6&365&1045&45&987.77\\
MOT17-11-FRCNN &77.20 &61.30&81.40&85.90&43&6&397&1701&46& 989.27\\
MOT17-11-SDP&76.00 &61.40&83.95&85.90&43&6&269&1712&43&992.01\\
MOT17-13-DPM&53.50 &41.30&83.77&72.70&78&5&2731&1495&330&1054.97 \\
MOT17-13-FRCNN&54.90 &43.00&86.70&73.60&80&5&2174&1520&333&1055.24\\
MOT17-13-SDP &52.00&39.30&86.80&73.70&78&5&1480&1528&327&1057.83\\ \hline
\textbf{OVERALL}&\textbf{67.20}&\textbf{58.50}&\textbf{94.30}&\textbf{86.80}&\textbf{859}&\textbf{164}&\textbf{13763}&\textbf{24878}&\textbf{3297}&\textbf{15795.79}\\ 

\end{tabular}
\end{adjustbox}
    \caption{Tracktor tracking results for MOT17 for private detection with reid, and motion}
    \label{tracktor+++3}
\end{table}
\subsection{Discussion of results}
Following the experimental studies, this section provides a discussion on the performance of the two tracking algorithms (SORT and Tracktor++). The performance evaluation criteria were based on the metrics that were identified in Section 6. Both algorithms were evaluated using the MOT17 training dataset.
Table ~\ref{sort-tracking} presents the tracking results of SORT on the MOT17 dataset. Likewise Tables \ref{tracktor++1}, \ref{tracktor++2},  and \ref{tracktor+++3} present the tracking results of Tracktor++ on the same dataset. Comparing the tracking performance of Tracktor++ without additives (Motion, ReID) with SORT shows that Tracktor++ has superior performance over SORT. The multiple object tracking accuracy(MOTA), recall(Rcll), precision (Prcn), number of mostly tracked trajectories( MT) for Tracktor++ were 64.90\%, 55.70\%, 72.00\%,  95.70\%,  and 780 respectively as against 32.03\%, 23\%, 95.30\%,  and 126 for SORT.
Also, comparing the bare-bone Tracktor++ algorithm with Tracktor++ combined with the re-identification model, we saw as in  table \ref{tracktor++1} and  table \ref{tracktor++2} that there was a slight improvement in the MOTA, recall, and precision scores of Tracktor++ when it was augmented with the Re-ID model. Also, the number of tracklet ID switches is reduced when using Re-ID because it helps to recover tracklets that get missing temporarily  due to occlusions.
Furthermore, when, we activated both the motion model and re-identification network in Tracktor++, we saw as from the results in table \ref{tracktor+++3} that, there was an improvement in the performance of Tracktor++.
Thus, tradeoffs have to be made in multi-people tracking methods about which models to adopt in the tracking pipelines. The inclusion of motion and re-id models in the tracking pipeline improves the performance of Tracktor++ but, that comes at an increased cost in computation time as shown in Table ~\ref{tracktor++2}, and ~\ref{tracktor+++3}.

\section{Recommendation for future work}
Multi-object tracking is still an active research topic and the computer vision community has been making some significant progress in it. From the results of these two tracking algorithms, we can see that they are far from being the state of art algorithms. Specifically for the MOT Challenge, a lot of new algorithms have been proposed and they have better results than these two algorithms. However, from the analysis of these two algorithms, we can gain some insights about what could be done and possible techniques we could apply to improve the performance of multi-object tracking models.
First, we have identified that the quality of detections from the object detector has a great impact on the performance of the tracking algorithm. This is because if we have good detection results which we provide as input to the tracking pipeline, we can reduce the number of false alarms and also reduce the false negative scores of tracking algorithms.
Second, one of the major reasons for the poor performance of tracking algorithms is the presence of occlusions and crowded scenes. While recent research has pushed forward the quality of object detectors and in our case pedestrian object detectors to be precise, these detectors are still not immune to the effects of occlusions. As a result, the state-of-the-art object detectors, still do not perform well when predicting bounding boxes in crowded scenes. One of the ways of mitigating this problem is to provide more datasets containing crowded scenes which could be used to train detectors to offset this current problem.
Also, improving upon the techniques used to recover tracklets that  go missing for some time due to occlusions in the tracking pipeline would lead to some performance gains.
Moreover, the context contains useful information that could be exploited to improve the performance of multi-object tracking models. For instance, if we are able to capture the knowledge of human motion in tracking algorithms we could be able to improve the number of tracked trajectories (MT) of our tracking algorithms. However, many of the approaches so far have modeled human motion based on constant velocity assumptions. Adopting a motion model where we capture the dynamic motion of people could help us to recover lost tracklets. For instance, adopting a Long Short-Term Memory (LSTM) recurrent neural network to model the motion activities of humans in a tracking scenario would help us reduce the number of tracklets we lose due to occlusions.
Furthermore, Re-ID models are for recovering lost or missing tracks that occur due to occlusion. But, the performance of the Re-ID network has a great impact on the number of lost tracklets we can recover when they go missing due to noise. Thus,  using robust and high-performing Re-ID models could push forward the performance of tracking algorithms. Also, the RE-ID models adopted must be trained using data that is significantly related to the tracking problem otherwise, we could have a high number of false re-identifications which could reduce the performance of  tracking algorithms.
\section{Conclusion}
This paper presented a comparative study of two multi-person tracking algorithms. We have shown that the SORT algorithm is simple, but does not perform so well on the tracking task as the Tracktor++ algorithm does. We have also considered some trade-offs researchers have to make when adopting motion models. 
Also, tracking-by-detection is the common approach to the task of multiple-object tracking. We thus recommend that time should be spent on training and tuning object detectors to have good performance so that quality bounding boxes can be fed as input into the tracking algorithms we design. 
The usual assumption of modeling human motion using constant velocity heuristics is not always true. Thus, we argue that adopting models that try to handle the dynamic nature of human motion could improve the number of mostly tracked trajectories (MT) and minimize the number of ID switches (ID switches)  in tracking algorithms. 
Finally, the matching strategies based on object appearance similarity scores should use good object appearance  models that are trained using tracking relevant datasets to reduce the false recovery of tracklets as this  hurts the performance of tracking algorithms.


\clearpage

\bibliography{lipics-v2019-sample-article}

\end{document}